\RequirePackage[T1]{fontenc}
\documentclass[letterpaper,10pt,conference]{ieeeconf}
\IEEEoverridecommandlockouts
\usepackage{tikz}
\usetikzlibrary{arrows.meta,positioning,fit,backgrounds,calc,shapes.geometric,decorations.pathreplacing}
\usepackage{graphicx}
\usepackage{amsmath,amssymb}
\usepackage{booktabs}
\usepackage{multirow}
\usepackage{array}
\usepackage{xcolor}

\definecolor{cpteal}{RGB}{29,158,117}
\definecolor{cpblue}{RGB}{52,101,164}
\definecolor{cpsand}{RGB}{214,152,64}
\definecolor{cpgrey}{RGB}{90,96,104}

\usepackage{colortbl}
\usepackage[hidelinks]{hyperref}

\graphicspath{{fig/}{./}{figures/}}

\title{CoralPlan: Observation Skill Selection and Execution\\
for Underwater Robotic Inspection}

\author{Yuer Gao, Yu Zhao, and Yi Cai$^{*}$%
\thanks{All authors are with The Hong Kong University of Science and Technology (HKUST). $^{*}$Corresponding author: Yi Cai.}}

\begin{document}
\maketitle

\begin{abstract}
Underwater robotic inspection depends on acquiring views that reveal
task-relevant structure. For a structurally complex coral colony, recognising
the target is only the starting point: the robot must select and execute a
viewing motion suited to the inspection task. We present CoralPlan, a
vision--language system that selects an observation skill from a current
camera image and task text supplied by an episode manifest. A shared motion
interface executes orbit, patch, or survey as target-relative trajectories;
the remaining plan fields provide operator guidance. Observation completion
requires target keeping and primitive-specific coverage, while joint success
also requires selection to match the recorded reference. We evaluate this
interface in 144 simulated episodes and 36 matched simulation--hardware pairs.
In a clear-water pool with external target-reference poses, hardware
observation completion reaches 77.8\% and joint success reaches 63.9\%.
The experiments identify both reference-mismatched completions and incomplete
observations after a matching skill selection. These results connect
observation-skill choice to measurable underwater execution outcomes and
identify where task-directed acquisition succeeds or fails.

\end{abstract}

\begin{keywords}
Marine robotics, underwater inspection, robot sensing and perception,
vision--language models.
\end{keywords}

\section{Introduction}

Coral reefs face increasing environmental
threats~\cite{hoegh2007coral}, motivating repeatable robotic monitoring.
For a robot inspecting a colony, useful imagery is an outcome of its
motion: viewpoint, standoff, and viewing extent determine which features
are acquired. Recognising a target provides the starting context;
inspection then requires the robot to obtain observations suited to the
operator's goal. This acquisition decision connects underwater perception
to purposeful physical action. We study it as an observation-acquisition
problem: the robot's behaviour determines the sensory evidence acquired
for its task.

Corals provide a demanding setting for this problem. Their complex
three-dimensional forms create occlusions, while local appearance and
surface detail motivate observations at different distances and angles.
Reef photogrammetry depends on survey coverage and overlapping views to
recover structural detail~\cite{bayley2020protocol}. A close-range
condition check, a multi-angle examination, and a wider survey of the same
colony therefore call for different camera--target geometries.
These viewing requirements make observation planning an enabling component
of detailed assessment, three-dimensional reconstruction, and systematic
scanning. The robot must decide how to acquire the relevant evidence as
well as interpret the images already available to it.

\begin{figure}[t]
\centering
\includegraphics[width=\columnwidth]{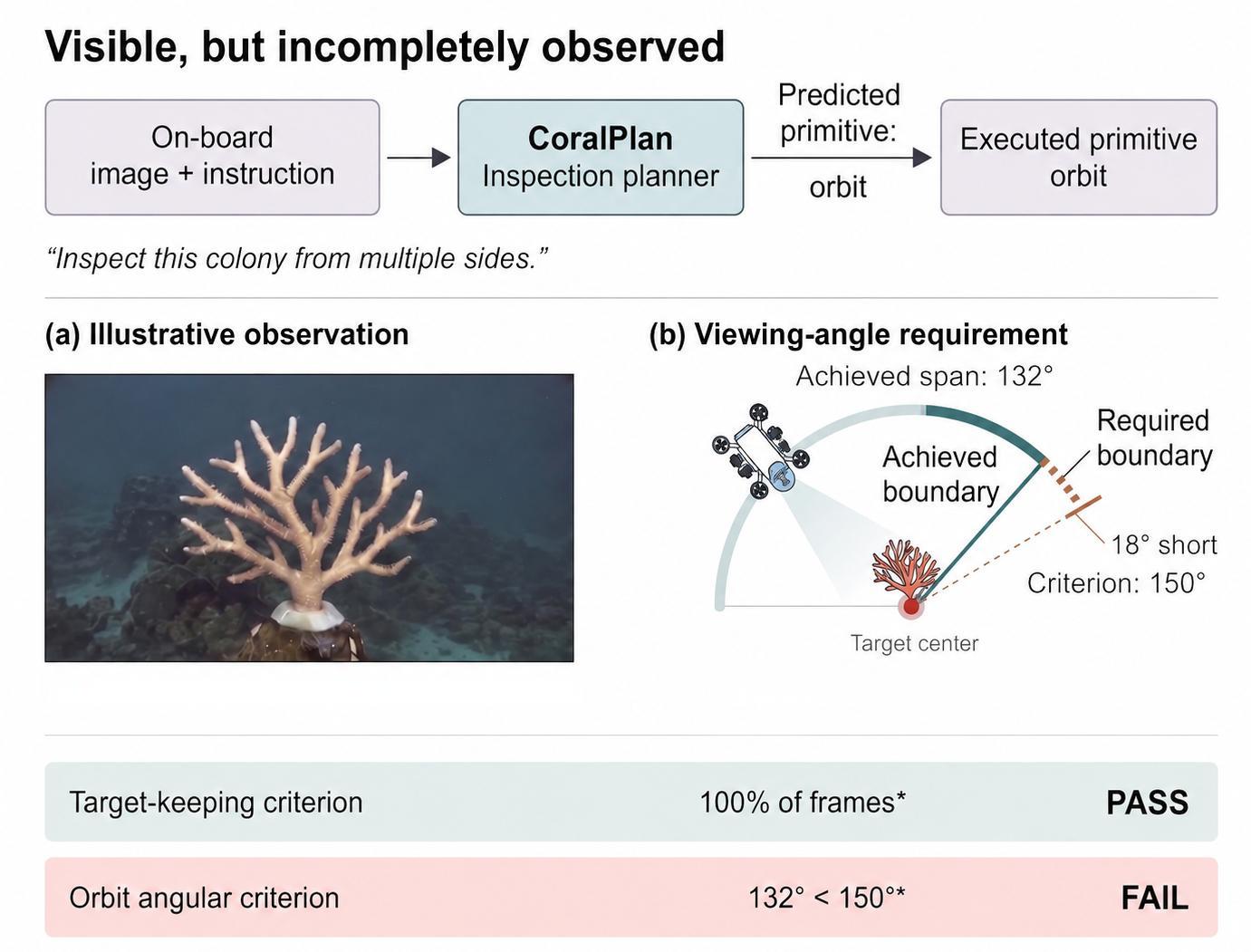}
\caption{\textbf{Visible, but incompletely observed.} CoralPlan selects
\emph{orbit} for multi-angle observation. The reported
episode passes target keeping ($K=1$) but attains a $132^\circ$
target-relative sweep, $18^\circ$ short of the $150^\circ$ criterion.
Asterisks mark episode measurements.}
\label{fig:teaser}
\end{figure}

Coral datasets support benthic annotation, condition recognition, and
scene segmentation~\cite{beijbom2015coralnet,shao2024coral,sauder2025coralscapes}.
Active perception and viewpoint planning address task-directed sensing and
coverage~\cite{bajcsy1988active,hollinger2013active,guedon2022scone}, while
language-guided underwater planners connect operator commands to navigation
and survey missions~\cite{oceanchat2023,oceanplan2024}.
We study the target-level interface between these capabilities: selecting
an observation strategy from visual context, executing that strategy, and
evaluating the acquired viewing geometry. Offline image--instruction tests
and camera-driven physical trials examine this interface at different stages.
Fig.~\ref{fig:teaser} illustrates the need for this last step: the target
remains in view, yet the executed sweep falls short of the
multi-angle observation criterion.

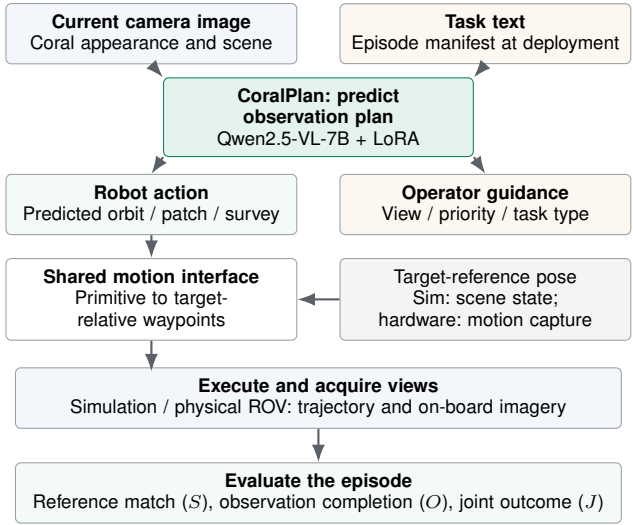
\begin{figure}[t]
\centering
\resizebox{\columnwidth}{!}{%
\begin{tikzpicture}[x=1cm,y=1cm,font=\sffamily\scriptsize,>=Latex,
 box/.style={draw=cpgrey!60,rounded corners=2pt,align=center,inner sep=4pt},
 flow/.style={->,thick,draw=cpgrey}]
\path[use as bounding box] (0,0) rectangle (8.6,6.9);
\node[box,fill=cpblue!6,text width=3.55cm] (image) at (2.1,6.45)
 {\textbf{Current camera image}\\Coral appearance and scene};
\node[box,fill=cpsand!8,text width=3.55cm] (cmd) at (6.5,6.45)
 {\textbf{Task text}\\Episode manifest at deployment};
\node[box,fill=cpteal!12,draw=cpteal!80!black,text width=3.8cm] (model) at (4.3,5.35)
 {\textbf{CoralPlan: predict observation plan}\\Qwen2.5-VL-7B + LoRA};
\draw[flow] (image.south)--(model.north west);
\draw[flow] (cmd.south)--(model.north east);
\node[box,fill=cpteal!5,text width=3.55cm] (primitive) at (2.1,4.2)
 {\textbf{Robot action}\\Predicted orbit / patch / survey};
\node[box,fill=cpsand!8,text width=3.55cm] (guidance) at (6.5,4.2)
 {\textbf{Operator guidance}\\View / priority / task type};
\draw[flow] (model.south west)--(primitive.north);
\draw[flow] (model.south east)--(guidance.north);
\node[box,text width=3.55cm] (map) at (2.1,2.95)
 {\textbf{Shared motion interface}\\Primitive to target-relative waypoints};
\node[box,fill=cpgrey!7,text width=3.55cm] (pose) at (6.5,2.95)
 {Target-reference pose\\Sim: scene state; hardware: motion capture};
\draw[flow] (primitive.south)--(map.north);
\draw[flow] (pose.west)--(map.east);
\node[box,fill=cpblue!6,text width=7.7cm] (exec) at (4.3,1.65)
 {\textbf{Execute and acquire views}\\Simulation / physical ROV: trajectory and on-board imagery};
\draw[flow] (map.south)--(2.1,2.2)--(2.1,2.0);
\node[box,fill=cpteal!5,text width=7.7cm] (eval) at (4.3,.4)
 {\textbf{Evaluate the episode}\\Reference match ($S$), observation completion ($O$), joint outcome ($J$)};
\draw[flow] (exec.south)--(eval.north);
\end{tikzpicture}}
\caption{From current imagery to subsequent observation. The predicted
primitive drives robot motion; the other plan fields guide the operator.
Target-reference poses instantiate the trajectory, and offline scoring
evaluates the completed episode. Physical inference uses the current
camera image together with task text supplied from the episode manifest.
Matched simulation and hardware episodes receive the same task text.}
\label{fig:pipeline}
\end{figure}

\textbf{CoralPlan} connects this acquisition decision to robot execution.
It predicts a structured plan centred on an observation primitive, and a
shared mapper instantiates orbit, patch, and survey as parameterised
trajectories. View, priority, and task type provide operator guidance.
\textbf{CoralVLM} supplies image--instruction--plan records for adaptation
and offline controls. The evaluation proceeds from input information to
physical outcomes: offline controls examine instruction and matched-image
contributions, 144 simulated episodes expose selection and observation
failures, and 36 matched simulation--hardware pairs test the same interface
and scoring definitions on a physical ROV (Fig.~\ref{fig:pipeline}).

\noindent\textbf{Contributions.}\par\nopagebreak[4]
\begin{enumerate}
\item \textbf{An executable observation-skill interface:} a current image
and task text drive skill selection, and a shared mapper instantiates the
selected orbit, patch, or survey as target-relative motion. Remaining plan
fields provide operator guidance. CoralVLM supports adaptation and offline
modality controls.
\item \textbf{Observation-oriented execution scoring:} target keeping and
primitive-specific coverage define observation completion. Selection
agreement, observation completion, and joint success use the same
definitions in simulation and on hardware.
\item \textbf{System experiments and failure analysis:} factorial simulation
and 36 matched simulation--hardware pairs locate failures between skill
selection and observation acquisition. In simulation, reference-matched
orbit episodes expose a multi-angle execution bottleneck; both cohorts
also contain completed reference-mismatched behaviours under the specified
acceptance criteria.
\end{enumerate}

\noindent\textbf{Scope.} CoralPlan's motion interface supports three fixed
observation skills. Deployment task text comes from each episode's manifest.
Target-relative waypoints use simulator scene state in simulation and
external motion capture on hardware. The clear-water pool study evaluates
four known printed coral models, whose separately captured photographs are
included in training. Observation criteria test mask agreement and
primitive-specific trajectory geometry. The 87 complete wild-reef records
are reserved for offline evaluation.

The dataset, code, prompts, and scoring scripts are available in an
anonymised repository.\footnote{\url{https://anonymous.4open.science/r/CoralVLM-3EC9/}}

\section{Related Work}
\textbf{Underwater coral perception.} Image-based reef monitoring spans
benthic point annotation and expert variability~\cite{beijbom2015coralnet},
multi-label condition recognition~\cite{shao2024coral}, and dense scene
segmentation~\cite{sauder2025coralscapes}, with robotic surveys feeding
automated analysis~\cite{manderson2018robotic}. These works establish what
can be inferred from acquired imagery. CoralPlan addresses which observation
behaviour to execute and whether execution meets its viewing requirements.

\textbf{Language-guided underwater planning.} Existing systems map commands
to mission-level navigation and survey
plans~\cite{oceanchat2023,oceanplan2024}. CORAL~\cite{wu2026coral} combines VLM
guidance with dynamics-based waypoint planning, while underwater robot
learning also addresses contact manipulation~\cite{aquabot2025}.
CoralPlan connects target-level inspection instructions to non-contact
observation strategies and evaluates target keeping and acquired coverage.

\textbf{Language-conditioned skill grounding.} Grounding instructions in
predefined skills~\cite{ahn2022saycan,liang2023codeaspolicies} and adapting
vision--language models to specialised tasks~\cite{li2023llavamed} are
established approaches. SayCan combines language relevance with skill
affordances~\cite{ahn2022saycan}, whereas Code as Policies generates
programs over perception and control APIs~\cite{liang2023codeaspolicies}.
CoralPlan grounds inspection instructions in an observation vocabulary
whose skills have explicit target-keeping and coverage criteria, connecting
behaviour selection to the views acquired during execution.

\textbf{Active perception and viewpoint planning.}
Active perception treats sensing as a task-directed process~\cite{bajcsy1988active};
viewpoint planners optimise exploration gain~\cite{bircher2016receding},
inspection uncertainty~\cite{hollinger2013active}, or learned
surface-coverage gain~\cite{guedon2022scone}.
CoralPlan brings operator-conditioned strategy selection and execution
evaluation into a shared observation interface. Viewpoint optimisers could
serve as lower-level executors within this interface.

\section{CoralVLM: Data for Observation Planning}
\label{sec:coralvlm}

CoralVLM supplies image--instruction--plan records for training and evaluating
CoralPlan, together with auxiliary coral-health annotations.

\begin{figure}[t]
  \centering
\includegraphics[width=\columnwidth]{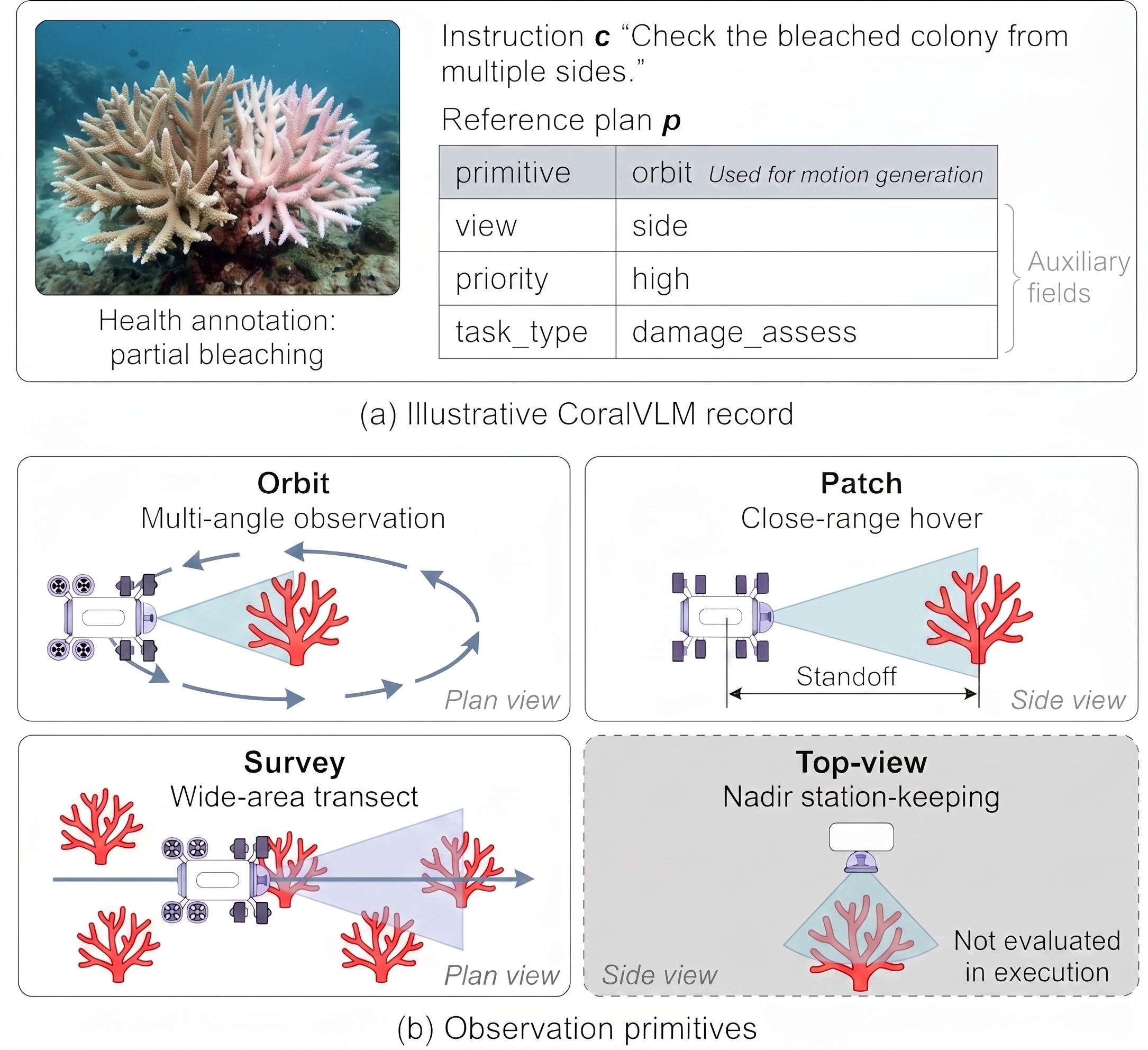}
  \caption{CoralVLM records and observation primitives.
  (a) An illustrative image--instruction--plan record with a health annotation.
  (b) Schematic viewing geometries, not to scale. Only the primitive field
  configures motion; top-view is in the planning vocabulary but is not
  supported by the current motion interface. Patch standoff is measured
  from the camera centre to the target reference point.}
  \label{fig:overview_primitives}
\end{figure}

\subsection{Record Definition}
Each record (Fig.~\ref{fig:overview_primitives}) contains an underwater image,
a health label (\emph{healthy}, \emph{partial bleaching}, \emph{severe
bleaching}, or \emph{dead}), an operator instruction, and a JSON plan.
CoralPlan predicts the plan. Health labels define the stratification used
to construct the balanced synthetic subset.

\subsection{Inspection Primitives}
\label{sec:primitives}
An inspection primitive specifies a camera--target observation geometry and
a trajectory family: close-range hover (\emph{patch}), multi-angle standoff
(\emph{orbit}), wide-area transect (\emph{survey}), or nadir station-keeping
(\emph{top-view}). This vocabulary links inspection goals to
vehicle-specific motion. For example,
``inspect the colony from multiple sides'' requests orbit, instantiated as
a target-relative arc. The current motion interface supports orbit, patch,
and survey; top-view is evaluated only at the planning level.

\subsection{Annotation Pipeline and Statistics}
GPT-4o first proposes health and primitive labels from an image, then uses
the image and primitive to generate an instruction and plan. Since both
derive from the same generated primitive, linguistic shortcuts are possible;
Table~\ref{tab:coralplan} evaluates modality contributions. Initial and
perturbed-prompt generation passes are pooled. A stratified 100-record human assessment gives
mean quality 4.4/5.0. Prompts and audit details are released.

The released set comprises 2{,}478 synthetic records
(460/928/494/596 healthy/partial/severe/dead) plus 100 real diver-captured
wild-reef images. Before scoring, the wild images were screened for
complete, reliable reference records. The 87 complete records have valid
annotations (40/36/2/9 healthy/partial/severe/dead) and form the held-out
wild planning set. The remaining 13 images were marked indeterminate and
excluded because they did not provide complete, reliable reference records
for the reported metrics.
Experiments use a class-balanced synthetic subset
of 1{,}792 records (448 per class) split 80/10/10 by stratified sampling
(seed 42) into 1{,}432 train, 180 validation and 180 synthetic-test records
(358/45/45 per class). Training also includes separately captured photographs
of the four physical coral models, distinct from the 87 wild evaluation
images. Physical deployment acquires new camera frames of these known
targets. The reported offline and execution models share this training
recipe.

Health balancing does not balance primitives. The synthetic test split is
31.1\% survey, 26.1\% top-view, 25.6\% orbit, and 17.2\% patch;
the majority primitive reference predicts survey.

We computed 64-bit dHash distances between the synthetic test split and
the synthetic development splits, covering 257{,}760 train--test and 32{,}400
validation--test pairs, for 290{,}160 pairs in total. The minimum Hamming
distance was 6; none of these pairs met the near-duplicate criterion of
distance $\leq5$.

\section{CoralPlan: From Intent to Observation Plans}
\label{sec:coralplan}

CoralPlan uses a vision--language model to select an observation plan; a
shared mapper instantiates its supported primitive as motion.

\subsection{Task Formulation}
CoralPlan predicts a JSON plan with \emph{primitive}, \emph{view},
\emph{priority}, and \emph{task type} slots. The primitive configures
motion; the remaining slots provide inspection guidance to the operator.
In offline evaluation, $(I,c)\rightarrow p$ pairs an underwater image $I$
with an inspection instruction $c$. At deployment, the program reads
task text from the episode manifest and supplies it with the current camera
frame. Matched simulation and hardware episodes use the same task text.

\subsection{Model and Adaptation}
CoralPlan adapts Qwen2.5-VL-7B~\cite{bai2025qwen25vl} to the
planning task using the synthetic training split and separate physical-target
photographs described in Section~\ref{sec:coralvlm}, with
LoRA~\cite{hu2022lora} on the language-model side, the vision encoder frozen
(rank 16, $\alpha$ 32, learning rate $1{\times}10^{-4}$, 5 epochs; training seeds 42, 43, and 44;
full configuration in the released repository). The
image-only reference adaptation uses the identical recipe without the
instruction. Plans use the same constrained-JSON output schema.
The seed-42 CoralPlan checkpoint is used for all reported robot execution
trials; the additional training seeds are evaluated offline.

\textbf{Splits and seeds.} The data partition uses seed 42 and is fixed
across training seeds 42, 43, and 44. Each model is evaluated on the same
180 synthetic-test and 87 held-out wild records. Adapted settings are
reported as mean $\pm$ sample standard deviation across the three training
seeds, computed before rounding. Rule-based references are single-run results. For shuffled images, each trained checkpoint is evaluated
under three reassignments: we first average the three scores within each
seed, then report the mean and sample standard deviation across the three
seed-level means. The nine shuffle evaluations yield three seed-level means per split.
Nominal 95\% Wilson intervals use a binomial working model for execution
proportions; training-seed standard deviations summarise checkpoint variation.

\subsection{Output Parsing and Reference Adaptation}
\label{sec:parsing}
\label{sec:refadapt}
The parser first attempts to extract a fenced JSON block or a bare JSON
object, then falls back to per-field regular expressions. Predictions with
at least three recovered fields are accepted. A fixed, released synonym
map normalises values.
Execution additionally requires a supported primitive. Unsupported or
unparseable predictions are not remapped and count as non-executable
failures whenever execution is required. The image-only reference predicts
a plan without an instruction using the same adaptation.

\section{From Observation Semantics to Motion}
\label{sec:execution}

An episode converts a camera observation into robot motion and a new set
of views. The planner selects a primitive, the motion interface instantiates
its trajectory, and the controller executes it. Evaluation then checks the
selection against the recorded reference and measures the achieved target
keeping and coverage (Fig.~\ref{fig:pipeline}).

\subsection{Execution Environment}
The OceanSim~\cite{oceansim2025}/Isaac Sim~\cite{isaacsim} environment
matches the physical ROV's geometry, propulsion configuration, camera
placement and field of view, and pool workspace. It contains four coral
targets. Pairing matches the initial layout, manifest-supplied task text,
and scoring criteria. Both platforms use the shared primitive-to-motion interface,
but select the primitive separately from their respective on-board images.

\subsection{Predicted-Plan Execution Loop}
At the start of a physical episode, the on-board camera supplies a current
image of the target. The program combines this frame with manifest task text for
CoralPlan to predict a structured plan. The mapper converts its supported
primitive into a circular standoff path, close-range
hover, or wide-area transect. A waypoint controller executes the trajectory
while the camera acquires further views. The view, priority, and task-type
fields provide operator guidance alongside the executed primitive.
Target-relative waypoint instantiation uses simulator scene state in
simulation and external motion capture on hardware.

\subsection{Execution Trial Matrix}
We run a controlled factorial suite of
$4 \times 4 \times 3 \times 3 = 144$ episodes: four coral targets, four
turbidity levels, three lighting levels, and three reference inspection
primitives (orbit, patch, survey). Episodes are grouped by the
\emph{reference} primitive, so each group contains exactly 48 episodes
irrespective of what the planner predicts. A supported prediction is executed
even when it differs from the reference, in which case the episode is scored
as a reference mismatch within its group. Unsupported or unparseable
predictions remain in that group and count as non-executable failures.
Three additional unlabelled recordings lie outside the factorial design.
All three are marked successful in the source log, but lack task labels;
they are excluded from the factorial and semantic analyses. Target keeping is the
fraction of evaluated frames with target-mask IoU $\geq 0.5$:
\[
K = \frac{1}{N}\sum_{t=1}^{N}
    \mathbf{1}\!\left[\mathrm{IoU}_t \geq 0.5\right].
\]
Mask agreement provides the target-keeping proxy.
Here, $\mathrm{IoU}_t$ compares the observed target mask
$M_t^{\mathrm{obs}}$ with the projected reference target mask
$M_t^{\mathrm{ref}}$ in the same camera frame. The projected reference
mask is clipped to the image bounds before IoU is computed:
\[
\mathrm{IoU}_t =
\frac{|M_t^{\mathrm{obs}}\cap M_t^{\mathrm{ref}}|}
     {|M_t^{\mathrm{obs}}\cup M_t^{\mathrm{ref}}|}.
\]
Empty or missing target masks are assigned $\mathrm{IoU}_t=0$ and
count as failures, including when both masks are empty. The denominator
$N$ includes all scored post-transient frames; only corrupted or
unsynchronized frames are excluded.
An episode satisfies the target-keeping criterion if $K \geq 0.8$.
Observation completion additionally requires the criterion for the
predicted, executed primitive:
\begin{itemize}
\item \textbf{Orbit}: target-relative azimuth span of at least
$150^\circ$. This threshold operationalises a broad multi-angle
inspection, distinguishing orbit from the local, single-viewpoint
observation of \emph{patch}.
\item \textbf{Patch:} camera--target distance $d_t$ satisfies
$|d_t-d_p^\ast|\leq\epsilon_p$ for at least $\rho_p=0.80$ of evaluated
frames, with $d_p^\ast=0.30\,\mathrm{m}$ and
$\epsilon_p=0.05\,\mathrm{m}$.
The distance is measured from the camera optical centre to the target
reference point.
\item \textbf{Survey:} target-relative camera position projected onto the
commanded survey axis, $s_t$, attains
$\max_t s_t-\min_t s_t>L_s=3.50\,\mathrm{m}$. Span, rather than raw
travelled distance, prevents oscillation from inflating coverage.
\end{itemize}
These operational thresholds were fixed before scoring the execution trials.

\subsection{Paired Physical Evaluation}
\label{sec:paired}
On the physical ROV, CoralPlan predicts from current on-board imagery
and task text supplied from the episode manifest. The shared mapper
executes a supported prediction automatically. Hardware trials use a clear-water pool under controlled
lighting. We run $4\times3\times3=36$ episodes: four targets, three reference primitives, and three repetitions,
each with a matched simulated counterpart. A 36-pair manifest links the
reference, predicted, and executed primitives to the corresponding episode
measurements. Both platforms execute their own predictions; pairs with
different actions remain in the evaluation. We score coverage separately
for each executed primitive using unrounded logged measurements, including
the strict survey-span inequality $>3.50\,\mathrm{m}$. Coverage pass is
computed independently of the $K\geq0.8$ target-keeping test.

\begin{figure*}[t]
\centering
\begin{minipage}[t]{0.61\textwidth}
\vspace{0pt}
\resizebox{\linewidth}{!}{%
\begin{tikzpicture}
\path[use as bounding box] (0,0) rectangle (13.8,11.94);
\clip (0,0) rectangle (13.8,11.94);
\node[anchor=south west,inner sep=0pt] at (0,0)
 {\includegraphics[width=21.48cm]{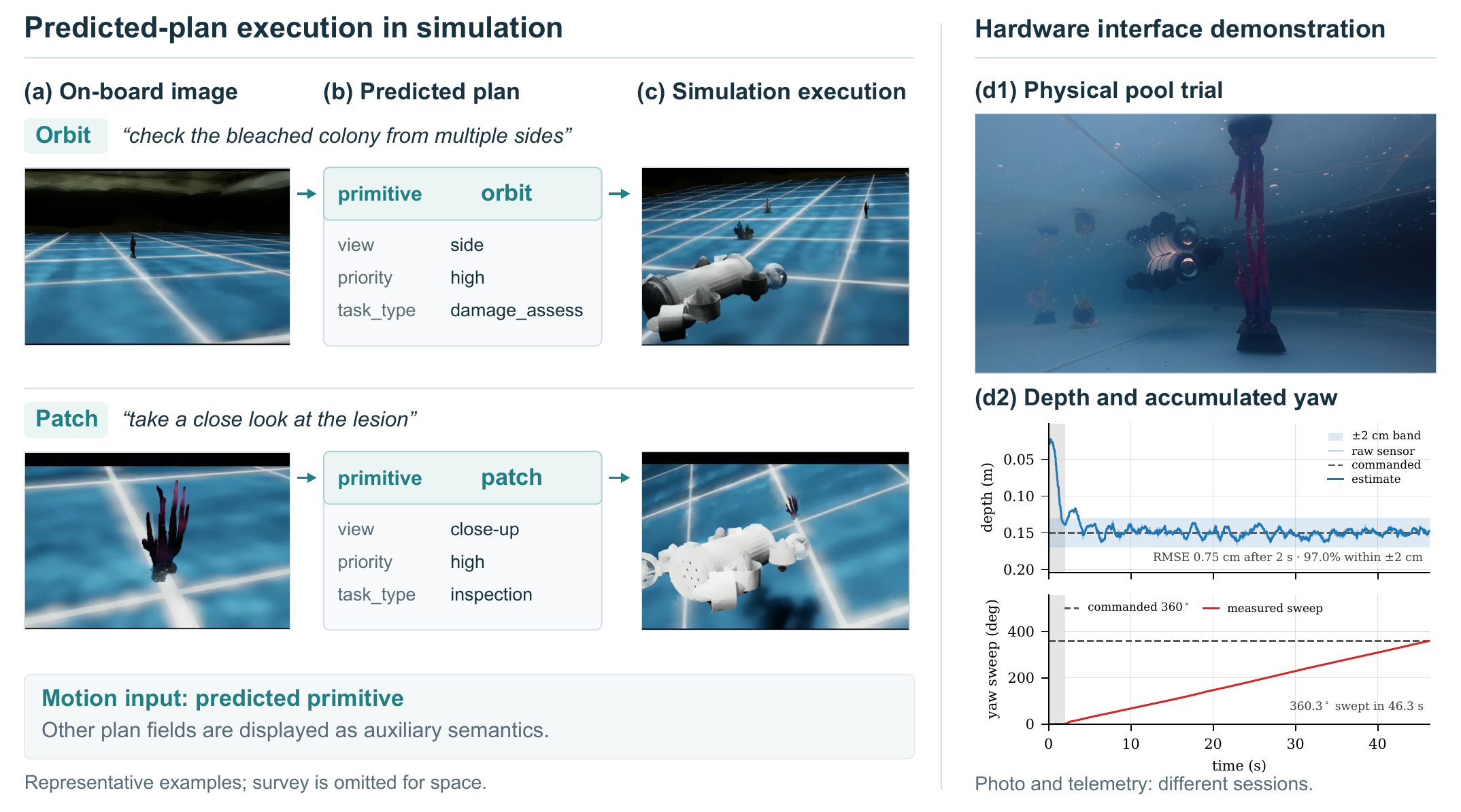}};
\end{tikzpicture}}
\end{minipage}\hfill
\begin{minipage}[t]{0.37\textwidth}
\vspace{0pt}
{\sffamily\small\bfseries (d) Physical pool trial}\par\smallskip
\resizebox{\linewidth}{!}{%
\begin{tikzpicture}
\path[use as bounding box] (0,0) rectangle (6.90,3.84);
\clip (0,0) rectangle (6.90,3.84);
\node[anchor=south west,inner sep=0pt] at (-14.33,-6.44)
 {\includegraphics[width=21.48cm]{predicted_execution_hardware.png}};
\end{tikzpicture}}
\par\medskip
\resizebox{\linewidth}{!}{%
\begin{tikzpicture}[x=1cm,y=1cm,font=\sffamily\scriptsize,
 cell/.style={draw=cpgrey!60,rounded corners=2pt,align=center,
 text width=2.75cm,minimum height=1.05cm,inner sep=3pt}]
\path[use as bounding box] (0,-.62) rectangle (6.4,3.58);
\node[anchor=west,font=\sffamily\bfseries\small] at (0,3.38)
 {(e) Target keeping and coverage};
\node[cell,fill=cpteal!10] at (1.55,2.50)
 {$K$ pass, $C$ pass\\\textbf{Sim 30 / HW 28}\\Observation complete};
\node[cell,fill=cpsand!13] at (4.85,2.50)
 {$K$ pass, $C$ fail\\\textbf{Sim 5 / HW 6}\\Insufficient coverage};
\node[cell,fill=cpblue!10] at (1.55,1.22)
 {$K$ fail, $C$ pass\\\textbf{Sim 1 / HW 0}\\Target-keeping failure};
\node[cell,fill=cpgrey!10] at (4.85,1.22)
 {$K$ fail, $C$ fail\\\textbf{Sim 0 / HW 2}\\Both criteria fail};
\node[align=left,anchor=north west,text width=6.25cm] at (0,.42)
 {Counts cover all 36 matched pairs.\\
 Observation completion requires both criteria.};
\end{tikzpicture}}
\end{minipage}
\caption{(a)--(c) On-board images, predicted plans, and simulated execution
for orbit and patch; only the primitive configures motion.
(d) Physical pool setup, shown as an interface illustration.
(e) Target-keeping and executed-primitive coverage verdicts across the
36-pair cohort, separating insufficient coverage from target-keeping
failures. Counts are reported separately for simulation and hardware.
The single coverage-only episode attains a $3.79\,\mathrm{m}$ survey span
with $K=0.72$.}
\label{fig:simreal}
\end{figure*}

Both platforms use the same IoU, frame-selection, and missing-mask rules.
Observed masks come from target-instance segmentation in simulation and
offline, manually verified image masks on hardware. Hardware reference
masks project registered target geometry using synchronized motion-capture
pose and calibrated camera intrinsics/extrinsics, with clipping to the
image bounds. Masks are used exclusively for offline scoring. Motion capture
supplies target-reference poses for waypoint instantiation and evaluation.
Fig.~\ref{fig:simreal}(d) illustrates the physical pool setup;
panel (e) reports the paired cohort's target-keeping/coverage outcomes.

\textbf{Episode outcomes.} Let $r$ be the recorded reference primitive
(the manifest's \texttt{requested\_primitive} field) and
$\hat r$ the prediction. Selection agreement is
$S=\mathbf{1}[\hat r=r]$. Let $C_{\hat r}$ indicate whether the
executed primitive meets its coverage criterion. Observation completion is
$O=\mathbf{1}[K\geq0.8]C_{\hat r}$, with $O=0$ for a non-executable
prediction. Joint success is $J=SO$.
We report all three outcomes for the 144-episode factorial suite and for
both platforms in the 36-pair evaluation. Across paired episodes, action
agreement compares primitive identities, whereas observation and joint
agreement compare the respective binary $O$ and $J$ verdicts.

\begin{table}[t]
\centering
\caption{Selection and observation outcomes in 36 matched simulation--hardware pairs.
Rates are percentages with counts in parentheses. Agreement denotes
matching binary verdicts for the indicated metric. Coverage is scored for
the primitive executed on each platform; joint success requires both
reference-matched selection and observation completion.}
\label{tab:paired}
\footnotesize
\setlength{\tabcolsep}{3.2pt}
\begin{tabular}{@{}lccc@{}}
\toprule
Outcome & Simulation & Physical ROV & Agreement \\
\midrule
Target keeping        & 97.2 (35/36) & 94.4 (34/36) & 97.2 (35/36) \\
Coverage pass         & 86.1 (31/36) & 77.8 (28/36) & 91.7 (33/36) \\
Selection ($S$)       & 80.6 (29/36) & 77.8 (28/36) & 91.7 (33/36) \\
Observation ($O$)     & 83.3 (30/36) & 77.8 (28/36) & 94.4 (34/36) \\
\textbf{Joint ($J$)}  & \textbf{69.4} (25/36) & \textbf{63.9} (23/36)
                     & 88.9 (32/36) \\
\midrule
\multicolumn{4}{@{}l}{\emph{Selection--observation decomposition (counts)}} \\
Selection & Observation & Simulation & Physical ROV \\
Matched   & Complete   & 25 & 23 \\
Matched   & Incomplete & 4  & 5  \\
Mismatched & Complete   & 5  & 5  \\
Mismatched & Incomplete & 2  & 3  \\
\midrule
\multicolumn{4}{@{}l}{\emph{Observation completion by target (sim / real)}} \\
Target 1 & \multicolumn{3}{c}{100.0 / 100.0} \\
Target 2 & \multicolumn{3}{c}{88.9 / 88.9} \\
Target 3 & \multicolumn{3}{c}{77.8 / 66.7} \\
Target 4 & \multicolumn{3}{c}{66.7 / 55.6} \\
\bottomrule
\end{tabular}
\end{table}

\section{Experiments}
\label{sec:exp}

\subsection{From a Camera Frame to the Next Observation}
A physical trial starts with the ROV viewing a coral target. The current
frame and program-supplied task text are sent to CoralPlan. Its predicted
primitive activates the shared motion interface, and the controller executes
the resulting waypoints.
An orbit changes the viewing angle around the target; a patch holds a
close viewing distance; a survey extends observation along a transect.
The operator receives the auxiliary plan fields as inspection guidance.
After execution, logged imagery and geometry determine target keeping and
coverage, and the predicted primitive is compared with the episode's
recorded reference (Fig.~\ref{fig:simreal}).

Three studies examine this sequence. \textbf{Offline planning} compares
model predictions on image--instruction records to measure modality use.
\textbf{Factorial simulation} executes predictions in 144 combinations of
target, turbidity, lighting, and reference primitive. \textbf{Paired
deployment} compares 36 physical trials with their simulated counterparts,
each platform selecting from its own camera image. The results below first
examine selection, then the observations obtained through execution.

\subsection{Offline Planning Measures Instruction and Image Contributions}

\textbf{Setup.} Models receive image--instruction records from the fixed
synthetic and wild test splits. Primitive accuracy scores the selected
skill; Exact requires all four plan slots to match, and Plan-F1 averages
their macro-F1 scores, excluding health. Comparisons use a majority reference, keyword rules,
instruction-only and image-only adaptations, shuffled images, and CoralPlan.
The image-only reference is LoRA-adapted without instructions.

\textbf{Instruction information.}
CoralPlan reaches $74.4\pm1.1\%$ synthetic-test primitive accuracy across
three training seeds. Table~\ref{tab:coralplan} contrasts this with 31.1\%
for majority prediction, 54.4\% for keyword rules, $64.8\pm1.4\%$ for
instruction-only adaptation, and $42.8\pm1.1\%$ for the image-only
reference. Instructions supply substantial selection information.
The next comparison tests matched images while preserving the
image--instruction input format.

\textbf{Matched visual context.}
The shuffled-image control reassigns images within a split while retaining
instructions and reference plans. Unlike modality removal, it preserves the
trained input format while disrupting image--instruction correspondence.

On the synthetic test split, CoralPlan reaches $74.4\pm1.1\%$ primitive
accuracy, compared with $67.3\pm0.8\%$ for shuffled images, a difference
of $7.1$\,pp between the displayed means. Exact-plan accuracies are
$50.9\pm0.8\%$ and $42.8\pm0.6\%$, respectively.
The controls share data partitions, schema, and parser.
The matched-image scores exceed the shuffled-image scores on this split.

\textbf{Wild-image evaluation.}
On the wild split, majority prediction achieves the highest primitive
accuracy, whereas CoralPlan achieves higher aggregate Plan-F1 than the
majority reference across the four plan fields (Table~\ref{tab:coralplan}).
The two metrics expose different aspects of performance on this
class-concentrated split.

\begin{table}[t]
\centering
\caption{Planning results and modality controls. Shaded rows show class-frequency
references.
Synthetic $n=180$; wild $n=87$. Prim and Exact are percentage accuracies.
Plan-F1 is on a 0--100 scale. Adapted settings report mean $\pm$ sample
standard deviation over three training seeds; rules are single-run results.
For shuffled images$^{\dagger}$, each seed's score first averages three
shuffles. ``--'' denotes unreported metrics.}
\label{tab:coralplan}
\footnotesize
\setlength{\tabcolsep}{2pt}
\begin{tabular}{@{}llccc@{}}
\toprule
Setting & Split & Prim & Exact & Plan-F1 \\
\midrule
\multicolumn{5}{@{}l}{\emph{Class-frequency references}}\\
\rowcolor{black!6} Majority class & Synth & 31.1 & -- & 28.6 \\
\rowcolor{black!6} Majority class & Wild  & 85.1 & -- & 34.2 \\
\midrule
\multicolumn{5}{@{}l}{\emph{Single-modality and shortcut controls}}\\
Keyword-rule & Synth & 54.4 & -- & -- \\
Keyword-rule & Wild  & 48.3 & -- & -- \\
Instruction-only & Synth & $64.8\pm1.4$ & $42.2\pm1.1$ & $55.6\pm0.6$ \\
Instruction-only & Wild & $50.6\pm1.1$ & $32.2\pm1.1$ & $47.4\pm0.7$ \\
Image-only ref. & Synth & $42.8\pm1.1$ & -- & $41.8\pm0.9$ \\
Image-only ref. & Wild & $24.1\pm1.1$ & -- & $31.4\pm0.7$ \\
Shuffled image$^{\dagger}$ & Synth & $67.3\pm0.8$ & $42.8\pm0.6$ & $58.4\pm0.8$ \\
Shuffled image$^{\dagger}$ & Wild & $51.3\pm1.8$ & $34.5\pm1.1$ & $49.2\pm0.8$ \\
\midrule
\multicolumn{5}{@{}l}{\emph{Full model}}\\
\textbf{CoralPlan} & Synth & $\mathbf{74.4\pm1.1}$ & $\mathbf{50.9\pm0.8}$ & $\mathbf{67.6\pm0.8}$ \\
\textbf{CoralPlan} & Wild & $\mathbf{56.3\pm1.1}$ & $\mathbf{39.1\pm1.1}$ & $\mathbf{53.8\pm0.9}$ \\
\bottomrule
\end{tabular}
\end{table}

\subsection{Simulation Exposes the Difficulty of Multi-Angle Observation}
\label{sec:exec}
Multi-angle observation is the most difficult of the three behaviours
under the main criteria. All reference-matched patch and survey episodes
complete, whereas only $20/34$ ($58.8\%$) reference-matched orbit episodes
meet both target keeping and the $150^\circ$ sweep requirement. The other
14 orbit episodes fail observation completion. Fig.~\ref{fig:teaser}
illustrates the physical distinction: a retained target with an insufficient
viewing sweep.

Across all 144 episodes, selection matches the recorded reference in
$111/144$ ($77.1\%$), observation completes in $121/144$ ($84.0\%$),
and both hold in $97/144$ ($67.4\%$; Table~\ref{tab:exec}). Of the
reference-matched selections, $97/111$ ($87.4\%$) complete. The orbit
group's $25/48$ ($52.1\%$) observation completions include the 20 orbit
completions above and five completed non-orbit predictions. Offline
accuracy (74.4\%) covers four primitives, including top-view; execution
evaluates the three supported reference types.

The four-way counts, ordered as $(S{=}1,O{=}1)$, $(1,0)$, $(0,1)$,
and $(0,0)$, are $(20,14,5,9)$ for orbit, $(38,0,10,0)$ for patch,
and $(39,0,9,0)$ for survey: $(97,14,24,9)$ overall. Thus the
24-episode gap between $O$ and $J$ counts completed reference-mismatched behaviours,
while 14 reference-matched episodes fail observation completion.
Of those 24 reference-mismatched completions, 19 belong to the patch- and
survey groups (10 and 9), which saturate the main observation
criteria. The sensitivity analysis below quantifies how completion changes
with stricter acceptance settings.

Within the orbit group, completion ranges from $6$--$7/12$
across turbidity levels and $7$--$9/16$ across lighting levels, compared
with $0/12$--$12/12$ across targets, highlighting variation among the four
tested colonies.

\begin{table}[t]
\centering
\caption{Simulated outcomes grouped by reference primitive (48 episodes each). Observation completion ignores selection
agreement; Joint requires both. $O\mid S=1$ is the observation-completion
rate among episodes with reference-matched selections. Mean $K$ averages
episode-level target-keeping scores.
All rates are percentages.}
\label{tab:exec}
\footnotesize
\setlength{\tabcolsep}{2.4pt}
\begin{tabular}{@{}lcccccc@{}}
\toprule
 & & \multicolumn{2}{c}{Executor only} & \multicolumn{3}{c}{Semantic decomposition} \\
\cmidrule(lr){3-4}\cmidrule(lr){5-7}
Reference & Ep. & Obs. & Target & Select & $O\mid S=1$ & Joint \\
 & & compl. & (mean $K$) & match & & success \\
\midrule
Orbit   & 48 & 52.1 & 88.5 & 70.8 & 58.8 & \textbf{41.7} \\
Patch   & 48 & 100.0 & 99.9 & 79.2 & 100.0 & \textbf{79.2} \\
Survey  & 48 & 100.0 & 99.8 & 81.3 & 100.0 & \textbf{81.3} \\
\midrule
Overall & 144 & 84.0 & 96.1 & 77.1 & 87.4 & \textbf{67.4} \\
\bottomrule
\end{tabular}
\end{table}

\textbf{Threshold sensitivity.} We rescore all 144 frozen seed-42 episodes
without replanning or rerunning trajectories. Thresholds apply by
\emph{executed} primitive, including reference-mismatched predictions in
other reference groups; results are grouped by \emph{reference} primitive as in
Table~\ref{tab:exec}. Target keeping, frame selection, the orbit criterion,
and non-executable-failure handling remain fixed.
Table~\ref{tab:sensitivity} reports observation completion decreasing from
$121/144$ to $112/144$, $100/144$, and $71/144$ under progressively tighter
criteria. In this rescore, completion in the orbit group remains $25/48$;
completion in the patch group becomes $44/48$, $39/48$, and $22/48$, and
completion in the survey group becomes $43/48$, $36/48$, and $24/48$.
The five non-orbit completions in the orbit group comprise two
patch and three survey executions, with $K=0.946$--$0.991$. At
$\epsilon_p=0.02\,\mathrm{m}$, the patch hold ratios are 0.958 and 0.971,
both above $\rho_p=0.95$. The survey spans are 4.083, 4.214, and
$4.356\,\mathrm{m}$, all strictly above $4.00\,\mathrm{m}$. All five
remain complete under Strict and the nested milder settings; together
with the 20 unchanged orbit completions, they account for $25/48$.
Each denominator of 48 denotes a reference group.
Table~\ref{tab:sensitivity} measures the combined sensitivity to jointly
tightened $\rho_p$, $\epsilon_p$, and $L_s$.
Per-parameter grids are available in the anonymised repository (footnote~1).

\begin{table}[t]
\centering\footnotesize
\caption{Observation completion under jointly tightened patch and survey
criteria on the same 144 frozen episodes. Patch standoff remains 0.30 m;
$\epsilon_p$ and $L_s$ are in metres. Survey uses strict $>L_s$.
The orbit criterion remains $\geq150^\circ$ and target keeping $K\geq0.8$.}
\label{tab:sensitivity}
\setlength{\tabcolsep}{3.2pt}
\begin{tabular}{@{}lcccc@{}}
\toprule
Setting & $\rho_p$ & $\epsilon_p$ & $L_s$ & Overall $O$ \\
\midrule
Main & 0.80 & 0.05 & 3.50 & 121/144 (84.0\%) \\
Mild & 0.90 & 0.04 & 3.70 & 112/144 (77.8\%) \\
Medium & 0.90 & 0.03 & 3.80 & 100/144 (69.4\%) \\
Strict & 0.95 & 0.02 & 4.00 & 71/144 (49.3\%) \\
\bottomrule
\end{tabular}
\end{table}

\subsection{Hardware Demonstrates Selection and Observation Outcomes}
\label{sec:pairedresults}
The physical ROV exhibits two distinct outcomes: five trials complete a
behaviour different from the recorded reference, while another five select
the reference primitive but fail its observation criteria. Of the remaining
26 trials, 23 satisfy both conditions and three satisfy neither.
The five reference-mismatched completions account for the $13.9$\,pp gap
between hardware observation completion and joint success.

Table~\ref{tab:paired} summarises all 36 pairs. Selection agrees with the
reference in $29/36$ simulated episodes ($80.6\%$) and $28/36$ hardware
episodes ($77.8\%$). Observation completion reaches $30/36$ ($83.3\%$)
and $28/36$ ($77.8\%$), respectively, while joint success is $25/36$
($69.4\%$) and $23/36$ ($63.9\%$).
Nominal 95\%
Wilson intervals for hardware $O$ and $J$ are [61.9, 88.3]\%
and [47.6, 77.5]\%, respectively; the corresponding simulation
intervals are [68.1, 92.1]\% and [53.1, 82.0]\%.

Predicted and executed primitives each agree across platforms in $32/36$
pairs ($88.9\%$); execution follows the prediction in every episode.
In three differing pairs, simulation executes patch and hardware executes
survey; in the fourth, simulation executes survey and hardware executes
patch. All four have matching $O$ verdicts despite different actions. Overall observation agreement is
$34/36$ ($94.4\%$): 28 pairs complete on both platforms, two complete
only in simulation, none complete only on hardware, and
six fail on both. Joint verdicts agree in $32/36$ pairs ($88.9\%$).
Among the 32 action-matched
pairs, observation verdicts agree in $30/32$ ($93.8\%$).

Target keeping passes in $35/36$ simulated and $34/36$ hardware episodes.
The target-keeping/coverage decomposition in Fig.~\ref{fig:simreal}(e) is
$(30,5,1,0)$ in simulation and $(28,6,0,2)$ on hardware, ordered as both
pass, keeping only, coverage only, neither. Thus five simulated and six
hardware episodes retain the target but fail coverage. One simulated
episode (P021 in the paired manifest) illustrates the reverse outcome:
its executed survey spans $3.79\,\mathrm{m}$, exceeding $3.50\,\mathrm{m}$,
but $K=0.72$ fails target keeping. The criteria therefore identify
different failures within this cohort. Observation completion follows the
same ordering across the four targets. This paired cohort is reported
separately from the 144-episode factorial suite.

\section{Limitations and Outlook}
\label{sec:limitations}

The generated instructions and plans share a primitive reference; independent
expert annotations and counterfactual input pairs would strengthen the
assessment of grounding. The wild split contains 87 complete records with
a concentrated primitive distribution. On this split, primitive accuracy and
Plan-F1 describe complementary aspects of agreement with the reference plans.

Execution evaluates a single adapted planner, the seed-42 checkpoint;
keyword-rule and instruction-only controls are evaluated offline.
Comparing their observation and joint outcomes requires additional
execution trials for the skills they select. The pool cohort evaluates
known target instances under the conditions stated in Scope. Additional
geometries, visually estimated target poses, and field trials would extend
this evidence. Orbit-threshold sensitivity and target-specific failure
analysis would further resolve the simulation bottleneck.

The current criteria measure observation geometry. Evaluating downstream
reconstruction quality and diagnostic detail would connect skill execution
to the utility of the acquired imagery for reconstruction and adaptive
scanning.

\section{Conclusion}
CoralPlan connects a current underwater view and task text to an executable
observation skill, linking target perception to the acquisition of further
views. A shared mapper turns the selected primitive into target-relative
motion, and target-keeping and coverage criteria assess the observation.
In simulation, all 14 episodes with reference-matched selections but
incomplete observations select and execute orbit. Separately,
reference-mismatched completions account for the gap between observation
completion and joint success in both cohorts. Hardware reaches 77.8\%
observation completion and 63.9\% joint success in the paired pool study.
The interface and experiments make the transition from observation-skill
selection to physical acquisition measurable, providing a system basis
for task-directed underwater inspection.

\section*{Acknowledgments}
GPT-4o generated candidate labels, instructions, and plans for the
dataset-construction pipeline in Section~\ref{sec:coralvlm}, followed by
filtering and sampled human assessment. OpenAI Codex
assisted with language editing and structural revision. The authors
reviewed and take responsibility for all scientific claims, analyses, figures,
and final content.

\bibliographystyle{IEEEtran}
\bibliography{CoralPlan_v30_refs}  

\end{document}